\documentclass[utf8]{frontiersSCNS} 

\usepackage{url,hyperref,lineno,microtype,subcaption}
\usepackage[onehalfspacing]{setspace}
\usepackage{booktabs}
\usepackage{makecell}
\usepackage{multirow}
\usepackage{amsthm}
\usepackage{amsmath}
\usepackage{amssymb}
\usepackage{algorithm}
\usepackage{algpseudocode}

\newtheorem{definition}{Issue}[section]

\linenumbers

\def\keyFont{\fontsize{8}{11}\helveticabold }
\def\firstAuthorLast{Sample {et~al.}} 
\def\Authors{Andrea Cossu\,$^{1,2,*}$, Gabriele Graffieti\,$^{3}$, Lorenzo Pellegrini\,$^{3}$, Davide Maltoni\,$^{3}$,  Davide Bacciu\,$^{1}$, Antonio Carta\,$^{1}$ Vincenzo Lomonaco\,$^{1}$}
\def\Address{$^{1}$Pervasive AI Lab, Computer Science Department, University of Pisa, Pisa, Italy\\
$^{2}$Scuola Normale Superiore, Pisa, Italy\\$^{3}$ Biometric System \& Smart City Lab, Computer Science Department, University of Bologna, Bologna, Italy}
\def\corrAuthor{Computer Science Department, University of Pisa, Largo Pontecorvo, 3, 56127 Pisa}
\def\corrEmail{andrea.cossu@sns.it}

\begin{document}
\onecolumn
\firstpage{1}
\nolinenumbers

\title{Is Class-Incremental Enough for Continual Learning?} 

\author[\firstAuthorLast ]{\Authors} 
\address{} 
\correspondance{} 

\extraAuth{}

\maketitle


\begin{abstract}
\section{} 
The ability of a model to learn continually can be empirically assessed in different continual learning scenarios. Each scenario defines the constraints and the opportunities of the learning environment. Here, we challenge the current trend in the continual learning literature to experiment mainly on class-incremental scenarios, where classes present in one experience are never revisited. We posit that an excessive focus on this setting may be limiting for future research on continual learning, since class-incremental scenarios artificially exacerbate catastrophic forgetting, at the expense of other important objectives like forward transfer and computational efficiency. In many real-world environments, in fact, repetition of previously encountered concepts occurs naturally and contributes to softening the disruption of previous knowledge. \\
We advocate for a more in-depth study of alternative continual learning scenarios, in which repetition is integrated by design in the stream of incoming information. Starting from already existing proposals, we describe the advantages such \emph{class-incremental with repetition} scenarios could offer for a more comprehensive assessment of continual learning models.

\tiny
 \keyFont{ \section{Keywords:} continual-learning, lifelong-learning, catastrophic-forgetting, class-incremental, class-incremental-with-repetition} 
\end{abstract}

\section{Introduction}
Continual learning models learn from a stream of data produced by nonstationary, dynamic environments \citep{parisi2019, lesort2020}. Since the data distribution may drift at any time, continual learning violates the i.i.d assumption behind traditional machine learning training procedures, giving rise to problems like catastrophic forgetting of previous knowledge \citep{mccloskey1989catastrophic}. \\
The issues faced by a continual learning model are heavily influenced by the specific implementation of the general continual learning scenario introduced above. In the last years, the most popular scenarios all refer to an \emph{experience}-based way of learning in classification tasks \citep{vandeven2018a, delange2021}. In these scenarios, learning is broken down into a, possibly unbounded, stream of experiences $S = e_1,\ e_2,\ e_3,\ \hdots$, with abrupt and instantaneous drifts between one experience and the other \citep{lomonaco2021}. Each experience $e_i$ brings a set of data, together with optional additional knowledge, like a task label (usually, a scalar value) which helps to uniquely identify the distribution generating the current data \citep{lesort2020}. \\
The surge of interest in continual learning has been initially driven by its application to deep learning methodologies and mostly oriented towards supervised computer vision tasks, like object recognition from images \citep{li2016, rusu2016}. Naturally, one of the most intuitive procedures to convert available computer vision benchmarks into viable continual learning benchmarks consisted in the concatenation of multiple datasets to simulate drifts in the data distribution (one dataset per experience, as in the protocol used by \cite{li2016}). This immediately allowed to leverage the vast amount of existing computer vision benchmarks and to rapidly test new continual learning strategies on large-scale streams. The learning objective was to classify patterns by assuming to know from which dataset each pattern arrived (referred to as task-incremental learning scenario in the literature \citep{vandeven2018a}). This task label information simplifies the continual learning problem, since patterns from different datasets can be easily isolated by the model during both learning and inference. \\ 
After a period in which task-incremental has remained the most studied continual learning scenario, the attention of the community has now turned to class-incremental scenarios \citep{rebuffi2017}, where experiments are conducted on a single dataset, with patterns split by class and without any knowledge about the task label, neither during training nor during inference. \\
In the class-incremental setting each experience $e_i$ contains a training dataset $D_i = \{(x_j, y_j)\}_{j=1,...,M}$, where $x_j$ is the input pattern and $y_j$ is its target class. The peculiar characteristic of class-incremental scenarios is that they partition the target class space by assigning a disjoint set of classes to each experience. Formally, be $\mathcal{C} = \{c_k\}_{k=1,...,C}$ the set of all classes seen by the model during training and be $\mathcal{C}_i$ the subset of classes present in experience $e_i$, class-incremental scenarios satisfies the following condition:
\begin{equation} \label{eq:intersect} 
    \mathcal{C}_i \bigcap \mathcal{C}_j = \emptyset, \quad \forall i \neq j.
\end{equation}
We will refer to the constraint expressed by Equation \ref{eq:intersect} as the \emph{no repetition} constraint. It simply states that classes present in one experience are never seen again in future experiences or, likewise, that each class is present in one and only one experience.

In this work, we discuss the reasons behind the success recently achieved by the class-incremental scenario and we highlight some of the problems connected with its assumptions and requirements. In particular, we believe that class-incremental scenarios impose a strong focus on catastrophic forgetting while, on the contrary, in many real-world scenarios such effect is less pronounced due to a natural repetition of previously encountered patterns. Class-incremental scenarios, although very convenient for quickly setting up experiments, possibly narrow future research paths by overshadowing other objectives like forward transfer \citep{lopez-paz2017} and sample efficiency \citep{diaz-rodriguez2018}, fundamental requirements for the achievement of a true continual learning agent.\\
We do not advocate for the complete dismissal of the class-incremental scenario, which has been proven useful to spark the interest around continual learning and to foster the design of new solutions. Instead, we aim at promoting the usage of alternative continual learning scenarios, in which previously encountered patterns may be revisited in the future. We call this family of scenarios \emph{class-incremental with repetition} and we show that some examples are already present in the literature, even though still understudied. We believe that class-incremental with repetition scenarios constitute a promising direction for a more robust and thorough assessment of continual learners' performance.


\section{Class-incremental scenarios} \label{sec:ci}
The class-incremental learning scenario is nowadays very popular in the continual learning community. Its \emph{simplicity} and \emph{ease of use} greatly fostered new studies and efforts towards mitigating catastrophic forgetting, the main problem faced by models when learning in this setting. Other continual learning scenarios present in the literature include task and domain-incremental learning \citep{vandeven2018a}, task-free and data-incremental learning \citep{delange2020a, aljundi2019d} and online continual learning \citep{lopez-paz2017, aljundi2019b}. Even though these scenarios propose alternative learning settings, none of them really address and discuss the assumptions and constraints that characterize the class-incremental scenario. In Section \ref{sec:cir}, we will present few notable exceptions \citep{stojanov2019, lomonaco2020a, thai2021} that work on continual learning scenarios for classification based on repetition of previously seen classes.\\
We now turn the attention to some of the limitations caused by the \emph{no repetition} constraint of class-incremental.
\begin{definition} \label{is:real}
\text{Repetition occurs naturally in many real-world environments.}
\end{definition}
Class-incremental learning is not aligned to many applications in which repetition comes directly from the environment. Examples include robotic manipulation of multiple objects, prediction of financial trends, autonomous driving tasks, etc.
A learning agent exposed to a continuous stream of information should be able to incrementally acquire new knowledge, but also to forget unnecessary concepts and to prioritize learning based on some notion of importance. Not all the perceived information should be treated equally: if a certain pattern never occurs again, it may be useless to still pretend to predict it correctly. In fact, the statistical re-occurrence of concepts and their temporal relationship could be considered as important sources of information to help determine the importance of what the agent perceives \citep{maltoni2016, cossu2021}. It is very hard to discern what to forget and what concepts to reinforce if all the information is treated equally. \\
Learning in a compartmentalized fashion hinders many of the possible insights an agent may draw from the complexity of the environment, eventually limiting its possibility to create its world model suitable to the changing task it has to tackle.\\
Another important side effect of the \emph{no repetition} constraint is stated in 
\begin{definition} \label{is:forgetting}
\text{Lack of repetition induces large forgetting effects.}
\end{definition}
Focusing on catastrophic forgetting would not be inconvenient if real-world problems were actually aligned with the characteristics of the class-incremental scenario \citep{thai2021}. As expressed by Issue \ref{is:real} however, this is not the case. \\
Moreover, Issue \ref{is:forgetting} led to the generally accepted statement that \textit{replay strategies are the most effective strategies for continual learning} \citep{vandeven2020}. 
However, as we will show in Section \ref{sec:cir}, alternative scenarios may greatly reduce the replay advantage, whenever natural replay occurs in the environment. An excessive focus on catastrophic forgetting may also limit the scope of continual learning research, which instead involves many different problems and objectives like optimizing the learning experience for forward transfer and few-shot learning \citep{lopez-paz2017} or training continuously with limited memory and computational resources (as in edge devices) \citep{diaz-rodriguez2018}. While there are many works hinting at the fact that continual learning is not only about catastrophic forgetting \citep{thai2021, diaz-rodriguez2018}, the continual learning scenarios in which most of the research operates is still one in which the forgetting is by far the most pressing problem.
 
\section{Class-incremental with repetition scenarios} \label{sec:cir}
\begin{figure}[t]
    \centering
    \includegraphics[width=0.95\textwidth]{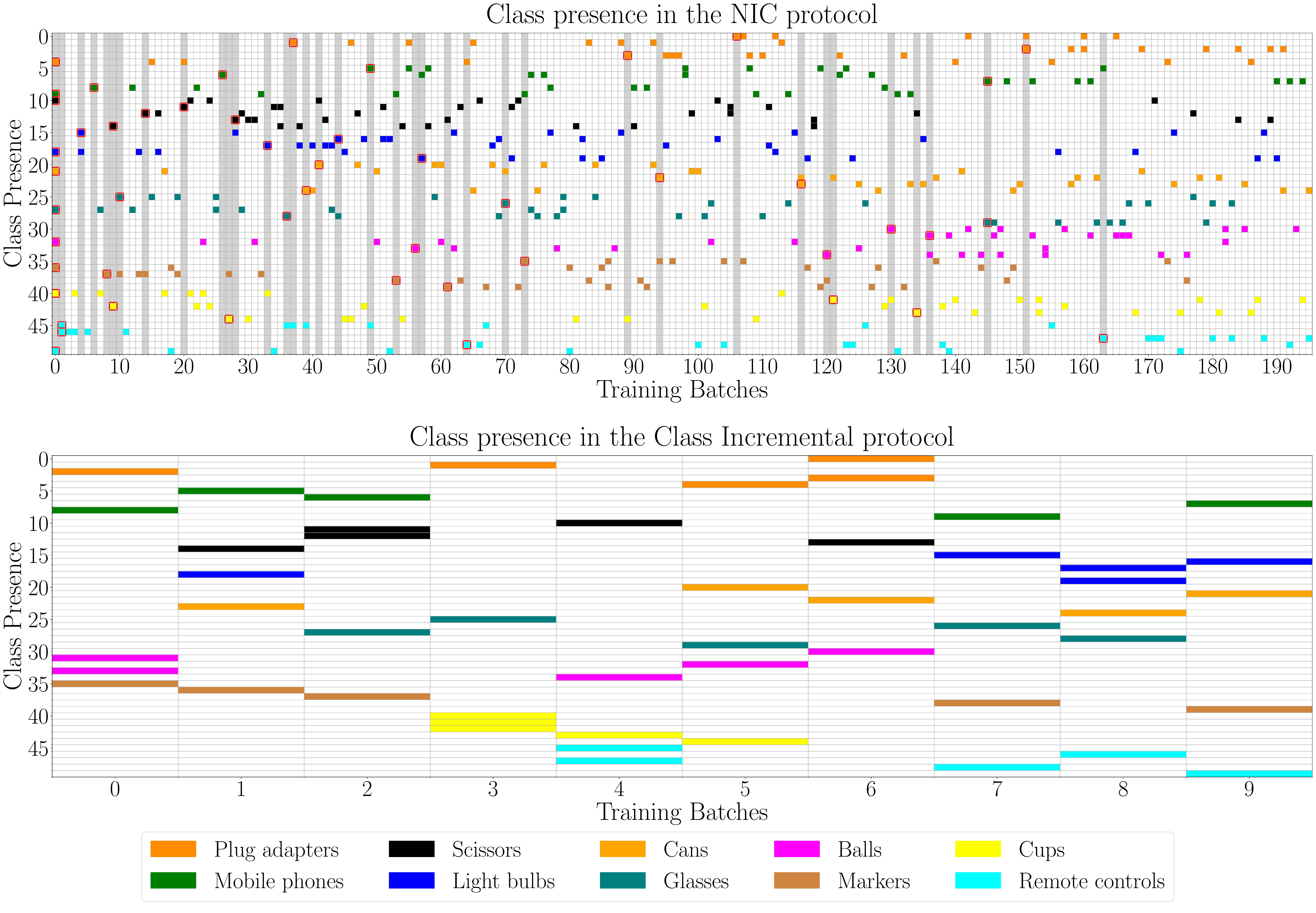}
    \caption{Comparison between class presence in continual learning streams from the NIC scenario (above) and class-incremental scenario (below). Each row represents a different class, while colors group classes into macro-categories (taken from CORe50 benchmark \citep{lomonaco2017}). The horizontal axis represents experiences (training batches). Gray vertical lines in the NIC scenario indicate the introduction of at least a new class in that experience (the newly introduced classes are surrounded by a red square). The NIC protocol shows a longer stream than class-incremental, with a more diverse distribution of the classes.}
    \label{fig:nic}
\end{figure}
The issues raised by the usage of class-incremental scenarios can be easily addressed by relaxing the \emph{no repetition} constraint. While there are many different ways of relaxing the constraint, the current literature already offers some proposals that can be grouped into the family of \emph{class-incremental with repetition} (CIR) scenarios.\\
An extreme example belonging to the family is the New Instances (NI) scenario, also called domain-incremental \cite{vandeven2018a, lomonaco2017, maltoni2019}. In NI, the first experience brings all the classes that the model will see in the subsequent experiences. Therefore new experiences only present new instances of previously seen classes. Depending on the amount of variation in the instances, the NI scenario may not present a large amount of forgetting, as in the popular Permuted MNIST benchmark. \\
\citet{stojanov2019} introduces the CRIB benchmark for incremental object learning based on exposure from contiguous view. Their incremental learning paradigm considers both the class-incremental scenario and the case of repetition of previous objects with different exposures (e.g., different 3D views). In the experiments, the authors found that allowing for even a small amount of repetition is beneficial to the continual performance, which approaches the one of joint training (offline) as more repetition is provided to the model. The results obtained by \citet{stojanov2019} were further confirmed also for unsupervised learning objectives, like reconstruction, in \citet{thai2021}. Their scenario with repeated exposures allowed to bridge the gap with the joint training performances without using any explicit replay of previous patterns, but only by leveraging the natural replay occurring in the environment.\\ 
The work done by \citet{lomonaco2020a} proposes a flexible setup for CIR (see Figure \ref{fig:nic} for a depiction of the resulting scenario). The authors based their experiments on the New Instances and Classes (NIC) continual learning scenario together with the CORe50 dataset, both introduced in \citet{lomonaco2017}. The NIC scenario is based on the assumption that each experience will bring patterns coming from both new and previously seen classes. Therefore, it fits well the CIR family of scenarios. Due to the property of the NIC scenario, \citet{lomonaco2020a} were able to experiment on a stream composed of a large number of experiences. The class-incremental counterpart produces a shorter stream since the total number of experiences is limited by the number of classes available in the dataset. The authors also provided the pseudocode of the protocol managing how many new classes will be introduced and in which experience. In contrast, the repeated exposure protocol of \citet{stojanov2019} and \citet{thai2021} only uses a random selection to sample which previous objects will be repeated. The design of a flexible procedure to generate CIR benchmarks is an important step since it can greatly contribute to foster the interest around the scenario itself. 

We now introduce a general protocol for build CIR data streams which, differently from the one by \citet{lomonaco2020a}, does not depend on the choice of a specific dataset. Instead, it can be instantiated on any dataset suitable for classification tasks. Algorithm \ref{alg:cir} presents the pseudocode for the protocol, which requires to specify the number of experiences and the classes in each experience (patterns are sampled at random from each class). The number of patterns per class in each experience is computed by dividing the available number of patterns for that class by the number of occurrences of the class in the stream of experiences (e.g., if class $c_0$ has $100$ patterns and is present in $4$ experiences, each experience will bring $25$ patterns of that class, selected at random without replacement from all the available ones). This is a simple protocol that can be customized in many ways: for example, by providing a custom selection policy for the patterns or by generating the classes in each experience with a properly designed algorithm. \\
The contributions discussed above show that CIR is already an available and ready-to-use scenario for continual learning experiments. 

\begin{algorithm}
\caption{Protocol to build class-incremental with repetition benchmark from existing classification dataset}
\begin{algorithmic}[1]
\Require Dataset $D = \{(x_j, y_j)\}_{j=1,...,M}$, number of experiences $N$ in the stream, list of classes per each experience $C = (\mathcal{C}_1, \mathcal{C}_2, ..., \mathcal{C}_N)$, where $\mathcal{C}_i = (c_i^1, c_i^2, ..., c_i^{k(i)})$ and $k(i)$ is the number of classes in experience $i$.
\item[]

\State $p \leftarrow [0, 0, ..., 0]$ with as many elements as number of classes in $D$. 
\For{$j \leftarrow 1, 2, ..., M$} \Comment{Number of patterns for each class in $D$}
    \State $p[y_j] = p[y_j]+1$
\EndFor
\State $e \leftarrow [0, 0, ..., 0]$ with as many elements as number of classes in $D$
\For{$i \leftarrow 1, 2, ..., N$} \Comment{Experience count for each class}
    \For{$k \leftarrow 1, 2, ..., k(i)$}
        \State $e[c_i^k] = e[c_i^k]+1$
    \EndFor
\EndFor
\State $p[i] \leftarrow floor(p[i] / e[i]),\ \forall i$
\item[]

\State \texttt{stream} $\leftarrow$ empty stream
\For{$i \leftarrow 1, 2, ..., N$} \Comment{build the stream of experiences}
    \State \texttt{stream}.append($E_i$)
    \State $\mathcal{C}_i \leftarrow C[i]$
    \For{$k \leftarrow 1, 2, ..., k(i)$}
        \State $c_i^k \leftarrow \mathcal{C}_i[k]$
        \State $P_i^k \leftarrow$ sample $p[i]$ patterns from $D$ belonging to class $c_i^k$
        \State $D$.remove($P_i^k$)
        \State $E_i$.add($P_i^k$)
    \EndFor
\EndFor
\State \Return \texttt{stream}
\end{algorithmic}
\label{alg:cir}
\end{algorithm}


\section{Discussion}
Continual learning is, admittedly, one of the grand challenges of artificial intelligence and machine learning. Being able to learn continuously from a progressive exposure to new concepts is fundamental for many real-world applications in which the environment is so diversified and complex that an offline training phase may never provide sufficient knowledge for the agent to succeed.\\
Continual learning research is still in its infancy. Therefore, it is only natural to rely on simplified experimental configurations and to design solutions that will not adapt to any possible situation. In this sense, we discussed the benefits that class-incremental scenarios brought to the study of continual learning, by providing a quick experimental configuration able to exploit the vast amount of existing resources, especially in the field of computer vision. Given the rapid prototyping opportunities and the large scale size of the datasets and benchmarks involved, class-incremental scenarios will still be useful for the continual learning community. \\
Nonetheless, it is important to realize that some assumptions behind class-incremental scenarios, the \emph{no repetition} constraint, in particular, are at the root of the issues we already described in Section \ref{sec:ci}. We presented a few alternative scenarios (Section \ref{sec:cir}) that are already present in the literature and that are better aligned with real-world environments in which repetition of previous concepts occurs naturally. Moreover, we provided an algorithm to build custom data streams for CIR scenarios which carefully balances the amount of new and previous knowledge seen by the model. The algorithm is particularly suitable for streams with a large number of small experiences (experiences carrying datasets with few patterns), a configuration present in many real-world applications and that cannot be modeled with a class-incremental scenario.
By putting less focus on catastrophic forgetting, CIR scenarios allow to better study alternative continual learning challenges, like exploiting existing representations to learn new information faster, or to identify which portion of knowledge should be kept intact and which portion may be forgotten. More in general, an environment equipped with repetition is an additional source of information that can be exploited by any continual learning agent during its lifetime.  \\
We believe these CIR scenarios are laying the foundations for a more diverse and thorough evaluation of the performance of a continual learning model, which better encompasses the principles and objectives of continual learning.

\section*{Conflict of Interest Statement}
The authors declare that the research was conducted in the absence of any commercial or financial relationships that could be construed as a potential conflict of interest.

\section*{Funding}
This work has been partially supported by the European Community H2020 programme under project TEACHING (Grant No. 871385).

\bibliographystyle{frontiersinSCNS_ENG_HUMS} 
\bibliography{CL}

\end{document}